\documentclass[10pt,twocolumn,letterpaper]{article}

\usepackage{iccv}              
\usepackage[accsupp]{axessibility}
\usepackage{array}
\usepackage{tabularx}
\usepackage{microtype}
\usepackage{multirow}
\usepackage{wrapfig}

\usepackage{pgfplots}
\pgfplotsset{compat=1.18} 
\usepackage{pgfplotstable}

\newcolumntype{P}[1]{>{\raggedright\arraybackslash}m{#1}}%
\newcolumntype{C}[1]{>{\centering\arraybackslash}m{#1}}%
\newcolumntype{R}[1]{>{\raggedleft\arraybackslash}m{#1}}%

\newcommand{\methodname}{GazeGaussian\xspace}

\definecolor{iccvblue}{rgb}{0.21,0.49,0.74}
\usepackage[pagebackref,breaklinks,colorlinks,allcolors=iccvblue]{hyperref}

\def\paperID{1336} 
\def\confName{ICCV}
\def\confYear{2025}

\title{GazeGaussian: High-Fidelity Gaze Redirection with 3D Gaussian Splatting}

\author{
    {\normalsize Xiaobao Wei$^{1,2}$ \quad Peng Chen$^{1,2}$ \quad Guangyu Li$^{1}$ \quad Ming Lu$^{3}$ \quad Hui Chen$^{1,2,\dagger}$ \quad Feng Tian$^{1,2}$}\\
    {\normalsize $^{1}$Institute of Software, Chinese Academy of Sciences}\\
    {\normalsize $^{2}$University of Chinese Academy of Sciences \quad $^{3}$Intel Labs China}\\
    {\normalsize weixiaobao0210@gmail.com}
}

\begin{document}
\maketitle
\begin{abstract}
Gaze estimation encounters generalization challenges when dealing with out-of-distribution data. To address this problem, recent methods use neural radiance fields (NeRF) to generate augmented data. However, existing methods based on NeRF are computationally expensive and lack facial details. 3D Gaussian Splatting (3DGS) has become the prevailing representation of neural fields. While 3DGS has been extensively examined in head avatars, it faces challenges with accurate gaze control and generalization across different subjects. In this work, we propose GazeGaussian, the first high-fidelity gaze redirection method that uses a two-stream 3DGS model to represent the face and eye regions separately. Leveraging the unstructured nature of 3DGS, we develop a novel representation of the eye for rigid eye rotation based on the target gaze direction. To enable synthesis generalization across various subjects, we integrate an expression-guided module to inject subject-specific information into the neural renderer. Comprehensive experiments show that GazeGaussian outperforms existing methods in rendering speed, gaze redirection accuracy, and facial synthesis across multiple datasets. The code is available at: \href{https://ucwxb.github.io/GazeGaussian}{https://ucwxb.github.io/GazeGaussian}. 
\end{abstract}   
\renewcommand{\thefootnote}{\fnsymbol{footnote}} 
\footnotetext[2]{Corresponding author.}
\section{Introduction}
\label{sec:intro}
\begin{figure}[!ht]
    \centering
    \includegraphics[width=0.90\linewidth]{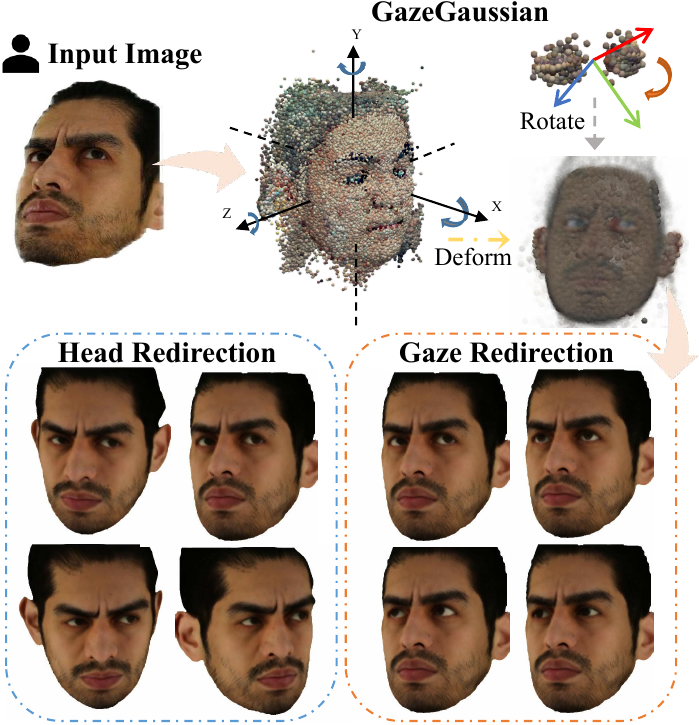}
    \caption{Gaze redirection: Given an input image, GazeGaussian deforms face and eye Gaussians from canonical space to generate high-fidelity head images with accurate gaze redirection.}
    \label{fig:intro}
\end{figure}
Gaze estimation is a fundamental component across various applications~\cite{andrist2014conversational, padmanaban2017optimizing, mavely2017eye}, yet current estimators~\cite{cheng2024appearance, cheng2022gaze, xu2023learning} often struggle to generalize effectively to out-of-distribution data. To address this, recent approaches~\cite{ruzzi2023gazenerf, yin2024Nerf_gaze, wang2023high} have started exploring gaze redirection, which manipulates the gaze in an input image toward a target direction. This process generates augmented data to enhance the generalization capabilities of gaze estimators.

Earlier methods~\cite{ganin2016deepwarp, yu2020unsupervised, zhang2022gazeonce, zhang18_etra} formulate gaze redirection as a 2D image manipulation task, relying on deep learning techniques to warp eye regions of the image toward the target gaze direction. However, these 2D approaches overlook the inherently 3D nature of head and gaze manipulation, often resulting in poor spatial consistency and limited synthesis fidelity. With advancements in Neural Radiance Fields (NeRF)~\cite{nerf} and its variants~\cite{wang2021neus, wei2024nto3d}, several methods~\cite{zielonka2023instant, grassal2022neural, zheng2023pointavatar, hong2022headnerf} have achieved 3D dynamic head representation and high-fidelity avatar synthesis. Meanwhile, to enable precise control of gaze direction, recent research~\cite{ruzzi2023gazenerf, yin2024Nerf_gaze, wang2023high} has introduced approaches that decouple the face and eye regions, modeling each with separate neural fields to achieve accurate gaze redirection.

As NeRF-based methods are hindered by high computational demands, 3D Gaussian Splatting~\cite{kerbl20233d} and its variants~\cite{lu2024scaffold, huang2024s3gaussian, wang2024plgs} achieve impressive rendering quality with significantly faster training speeds. Recent research~\cite{xiang2024flashavatar, qian2024gaussianavatars, xu2023gaussianheadavatar} has applied these methods to 3D head animation, typically using face-tracking~\cite{tran2018nonlinear, zielonka2022towards} parameters to model dynamic 3D head representations. However, existing 3DGS-based approaches neglect the accurate control of gaze direction and struggle to generalize across different subjects, limiting their effectiveness for gaze redirection.

To address the above issues, we propose GazeGaussian, a high-fidelity gaze redirection method that leverages a two-stream 3D Gaussian Splatting (3DGS) model to represent the face and eye regions, respectively. To our knowledge, this is the first integration of 3DGS into gaze redirection tasks. An overview is shown in Fig.~\ref{fig:intro}.

GazeGaussian begins by initializing the two-stream 3DGS model using a pre-trained neutral mesh on the training dataset. This mesh is divided into distinct regions for the face and eyes. By employing gaze direction and face tracking codes, we optimize a deformation field for the face and a rotation field for the eyes, allowing us to adjust the neutral Gaussians accordingly. To achieve precise eye rotation aligned with the target gaze, we present a novel Gaussian Eye Rotation Representation. In contrast to methods like GazeNeRF that implicitly alter feature maps, GazeGaussian explicitly adjusts the position of Gaussians in the eye branch according to the desired gaze direction, utilizing the controllable nature of 3DGS. 
To address possible errors in gaze direction, GazeGaussian develops an eye rotation field to enhance redirection accuracy. 
The two-stream Gaussians are rasterized into high-level features and sent to the neural renderer. 
Finally, to enable the generalization ability across different subjects and inject subject-specific information, we employ an expression-guided neural renderer (EGNR) to synthesize the final gaze-redirection images. 

Our main contributions are summarized as follows:
\begin{itemize}
\item We introduce GazeGaussian, the first 3DGS-based gaze redirection pipeline, achieving precise gaze manipulation and generalizable head avatar synthesis.

\item To enable rigid and accurate eye rotation based on the target gaze direction, we propose a novel two-stream 3DGS framework to decouple face and eye deformations, featuring a specialized Gaussian eye rotation for explicit control over eye movement.

\item To equip 3DGS with generalization ability, we design an expression-guided neural renderer (EGNR) to integrate subject-specific characteristics.

\item We conduct comprehensive experiments on ETH-XGaze, ColumbiaGaze, MPIIFaceGaze, and GazeCapture datasets, where GazeGaussian achieves state-of-the-art gaze redirection accuracy and facial synthesis quality with competitive rendering speed. 
\end{itemize}

\vspace{-2mm}
\section{Related Work}
\vspace{-1mm}
\noindent\textbf{Gaze Redirection.}
Gaze redirection is the task of manipulating the gaze direction of a face image to a target direction while preserving the subject's identity and other facial details.
Earlier approaches for gaze redirection include novel view synthesis~\cite{criminisi2003gaze,kuster2012gaze,giger2014gaze}, eye-replacement~\cite{qin2015eyegaze,shu2016eyeopener}, and warping-based methods~\cite{ganin2016deepwarp,kononenko2015gaze, gazedirector}. 
However, these methods are limited by person-specific data, restricted redirection range, and artifact introduction.
To further improve gaze redirection, recent studies~\cite{park2019few, selflearning, he2019gazeredirection, xia2020controllable} have employed neural network-based generative models. 
STED~\cite{selflearning}, building on FAZE~\cite{park2019few}, introduces a self-transforming encoder-decoder that generates full-face gaze redirection images.
With advancements in Neural Radiance Fields (NeRF)~\cite{nerf}, several studies~\cite{li2022eyenerf, ruzzi2023gazenerf, yin2024Nerf_gaze, wang2023high} have aimed to model the complex rotation of the eyeball. GazeNeRF~\cite{ruzzi2023gazenerf} employs a two-stream MLP architecture to separately model the face and eye regions, achieving improved gaze redirection performance.

However, these methods are hindered by substantial computational demands and limited rendering efficiency. Gaze manipulation occurs at the feature maps and remains an implicit approach. In contrast, GazeGaussian allows for explicit control over eye rotations, improving gaze redirection accuracy and accelerating the synthesis process.

\noindent\textbf{Head Avatar Synthesis.}
The synthesis of head avatars has garnered considerable attention in recent years.
FLAME~\cite{li2017learning} is a parameterized 3D head model that maps parameters of shape, expression, and pose onto 3D facial geometry, allowing for controllable head avatar generation.
Many subsequent works~\cite{VOCA2019,ranjan2018generating,fan2022faceformer,peng2023selftalk,peng2023emotalk,chen2023diffusiontalker} focus on using the FLAME model for speech-driven head avatar animation.
Recent head animation techniques can be categorized into two main approaches: NeRF-based methods and 3DGS-based methods. NeRF-based approaches~\cite{gafni2021dynamic, zielonka2023instant, hong2022headnerf, zheng2023pointavatar} leverage neural radiance fields to deform facial movements from a canonical space. 
HeadNeRF~\cite{hong2022headnerf} introduces a parametric head model that controls facial shape, expression, and albedo under different lighting conditions.
With the emergence of 3D Gaussian Splatting (3DGS)~\cite{kerbl20233d}, several approaches~\cite{qian2024gaussianavatars, xiang2024flashavatar, xu2023gaussianheadavatar, dhamo2023headgas} have explored its application in single head avatar modeling. Gaussian Head Avatar~\cite{xu2023gaussianheadavatar} initializes Gaussians with a neutral mesh head and incorporates MLPs to deform complex facial expressions. 

However, these methods ignore precise gaze control and are confined to the single avatar animation. In contrast, GazeGaussian achieves precise gaze direction control by decoupling facial animations and gaze movement within a two-stream model. Furthermore, an expression-guided neural renderer is designed to inject person-specific information for generalization ability across different subjects.

\vspace{-2mm}
\section{Overview}
\label{sec:overview}
\vspace{-1mm}
\begin{figure*}[!ht]

    \centering
    \includegraphics[width=0.90\textwidth]{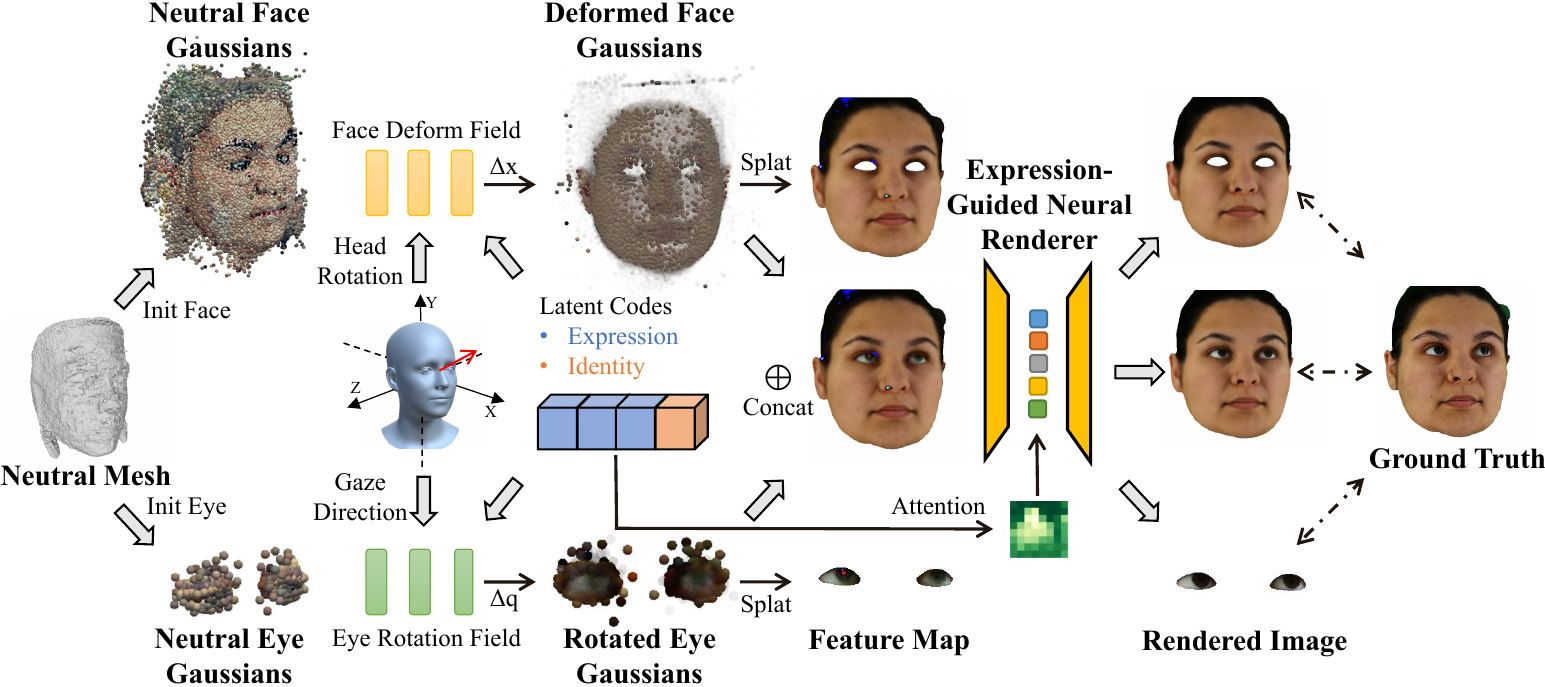}
    \caption{Pipeline of GazeGaussian. We initialize face-only and eye regions from a pre-trained neutral mesh. Using target expression codes, head rotation, and gaze direction, GazeGaussian optimizes face deformation and eye rotation fields to transform the neutral Gaussians. The transformed Gaussians are splatted into feature maps. The expression codes guide the neural renderer through cross-attention, enabling the rendering of feature maps into high-fidelity gaze redirection images, which are then supervised by multi-view ground truths.}
    \label{fig:pipeline}
\end{figure*}

The pipeline of GazeGaussian is illustrated in Fig.~\ref{fig:pipeline}, including the two-stream Gaussians and the proposed expression-guided neural renderer.
Before the beginning of the pipeline, we follow the data preprocessing in GazeNeRF~\cite{ruzzi2023gazenerf} and Gaussian Head Avatar~\cite{xu2023gaussianheadavatar}, which include background removal, gaze direction normalization, and facial tracking for each frame.
To obtain a neutral mesh for Gaussian initialization, we first reconstruct a Sign Distance Function (SDF) based neutral geometry and then optimize a face deformation field and an eye rotation field from the training data. A neutral mesh representing a coarse geometry across different subjects can be extracted using DMTet~\cite{shen2021deep}.
We then partition the neutral mesh into face-only and eye regions using 3D landmarks, initializing the two-stream Gaussians. Based on these neutral Gaussians, GazeGaussian optimizes a face deformation field and an eye rotation field to transform the Gaussians according to the target expression codes, gaze direction, and head rotation. Next, we concatenate the two-stream Gaussians and rasterize them into a high-dimensional feature map representing the head, face-only, and eye regions. Finally, these feature maps are fed into the expression-guided neural renderer to generate high-fidelity gaze redirection images. The ground truth image is used to supervise the rendered face-only, head, and eye images.

\vspace{-2mm}
\section{Method}
\vspace{-2mm}
\subsection{Preliminaries}
\vspace{-1mm}
The vanilla 3D Gaussians~\cite{kerbl20233d} with $n$ points are represented by their center $\mu \in \mathbb{R}^3$, the color computed by $k$ degrees of spherical harmonics $c \in \mathbb{R}^{3(k+1)^2}$, the quaternion rotation $R \in \mathbb{R}^{4} $, scale $S \in \mathbb{R}^{3} $ and opacity $\alpha \in \mathbb{R}$. The covariance matrix $\Sigma$ can be obtained by $\Sigma = RSS^{T}R^{T}$, which then formulates the Gaussian function $G(x) = e^{- \frac{1}{2}(x-\mu)^T\Sigma^{-1}(x-\mu)}$. These Gaussians are then rasterized into an image $I$, in which the pixel color $C$ is computed by blending $N$ ordered Gaussians overlapping the pixel: 
\begin{equation}
    C = \sum^N_{i=1}{c_i \alpha_i \prod_{j=1}^{i-1}}(1-\alpha_j),
\end{equation}
\subsection{Two-stream GazeGaussian Representation}
Our task is to synthesize a head avatar conditioned on gaze direction, head rotation, and expression latent codes. To decouple the complex movements in the face and eyes, we introduce a two-stream Gaussian model consisting of a face-only deformation branch and an eye rotation branch in the following section. 
\subsubsection{Face Deformation}
For the face-only branch, inspired by Gaussian Head Avatar~\cite{xu2023gaussianheadavatar}, we first construct canonical neutral face Gaussians with trainable attributes: $\{\mu_0^f, z_0^f, R_0^f, S_0^f, \alpha_0^f\}$. 
All attributes are identical to a standard Gaussian, except for $z_0^f \in \mathbb{R}^{128}$, which signifies the point-wise feature vectors. Rather than directly modeling complex color $c^f$, we decode $z_0^f$ into $c^f$ and affect other attributes' deformation to capture intricate facial details.

To transform canonical face Gaussians into the target space, we first model the influence of latent codes $\tau$ and head pose $\gamma$ through respective weights $\lambda_\tau$ and $\lambda_\gamma$.
We compute the minimum distance $d$ from each Gaussian center $\mu$ to the 3D facial landmarks (excluding eye regions). 
Based on the distance, we assign influence weights $\lambda_\tau$ and $\lambda_\gamma$ to the expressive latent codes $\tau$ and head pose $\gamma$. 
For Gaussians near landmarks ($d < d_1$), they are primarily influenced by expressive latent codes and set $\lambda_\tau = 1$. For distant Gaussians ($d > d_2$), they are predominantly affected by head pose, with $\lambda_\tau = 0$. To ensure a smooth transition between these regions ($d_2 \geq d \geq d_1$), we define $\lambda_\tau = (d_2 - d)/(d_2 - d_1)$. Consequently, the pose weight is determined as $\lambda_\gamma = 1 - \lambda_\tau$. The distance thresholds are empirically set to $d_1 = 0.15$ and $d_2 = 0.25$.

After that, we develop MLPs $\{E_\mu^f, E_c^f, E_\kappa^f\} \in E^f$ for latent codes-conditioned expression and $\{P_\mu^f, P_c^f, P_\kappa^f\} \in P^f$ for head pose-conditioned deformation:
\vspace{-2mm}
\begin{equation}
\begin{split}
 \mu^f &= \mu_0^f + \lambda_\tau E_\mu^f(\mu_0^f, \tau) + \lambda_\gamma P_\mu^f(\mu_0^f, \gamma), \\
 c^f &=  \lambda_\tau E_c^f(z_0^f, \tau) + \lambda_\gamma P_c^f(z_0^f, \gamma), \\
 \kappa^f &= \kappa_0^f + \lambda_\tau E_\kappa^f(z_0^f, \tau)
+\lambda_\gamma P_\kappa^f(z_0^f, \gamma),
\end{split}
\end{equation}
where $\kappa^f=\{R^f, S^f, \alpha^f\}$ and $\kappa_0^f=\{R_0^f, S_0^f, \alpha_0^f\}$. Finally, we apply rigid rotations and translations to transform Gaussians in the canonical space to the world space. Then, these Gaussians are rasterized into the feature maps:
\vspace{-2mm}
\begin{equation}
\mathcal{M}_f = \mathcal{R}(\{\mu^f, c^f, R^f, S^f, \alpha^f\}),
\end{equation}
where $\mathcal{R}$ represents the rasterizer and $\mathcal{M}_f$ indicates the feature map from the face-only branch.

\subsubsection{Eye Rotation}
Different from Gaussian Head Avatar~\cite{xu2023gaussianheadavatar}, we design an eye branch for accurate eye modeling and gaze redirection. 
For the eye branch, we construct canonical neutral eye Gaussians with attributes $\{\mu_0^e, z_0^e, R_0^e, S_0^e, \alpha_0^e\}$. 
These attributes share the same dimension as those in the face-only branch, except that $S_0^e \in \mathbb{R}^{N \times 1}$ is constrained to be spherical, aligning with the rotational properties of the eyeball. Since eyes are relatively small and mainly influenced by the gaze direction, $\lambda$ used in the face is omitted here. 

Directly applying the same face deformation strategy fails to fully leverage the unique characteristics of eyeball rotational motion, resulting in inaccurate gaze redirection. 
Therefore, we first rotate the eye Gaussians in the canonical space and then incorporate the expressive latent codes $\tau$ to generate deformation offsets. 
Since the gaze labels contain noise, directly using the normalized gaze direction $\varphi$ to rotate the Gaussians would lead to numerical optimization errors. To address this, we optimize two separate MLPs: $E_\mu^e \in E^e$ and $G_\mu^e \in G^e$ to predict the biases for Gaussian eye rotation.
\vspace{-2mm}
\begin{equation}
\begin{split}
 \mu^e = E_\mu^e(\mu_0^e, \tau) + G_\mu^e(\mu_0^e, \varphi) \mu_0^e,
\end{split}
\end{equation}
Other attributes are deformed in a similar way with $\{E_c^e,E_a^e\} \in E^e$ and $\{G_c^e,G_a^e\} \in G^e$:
\vspace{-2mm}
\begin{equation}
\begin{split}
c^e &= E_c^e(z_0^e, \tau) + G_c^e(z_0^e, \varphi), \\
\kappa^e &= \kappa_0^e + E_\kappa^e(z_0^e, \tau) + G_\kappa^e(z_0^e, \varphi),
 \end{split}
\end{equation}
where $\kappa^e=\{R^e, S^e, \alpha^e\}$ and $\kappa_0^e=\{R_0^e, S_0^e, \alpha_0^e\}$. By the proposed explicit eye rotation, GazeGaussian succeeds in face and eye disentanglement, which is ignored in the Gaussian Head Avatar~\cite{xu2023gaussianheadavatar}. Finally, eye Gaussians are transformed into the world space. Then these eye Gaussians are rasterized into the feature maps:
\vspace{-2mm}
\begin{equation}
    \mathcal{M}_e = \mathcal{R}(\{\mu^e, c^e, R^e, S^e, \alpha^e\}),
\end{equation}
To obtain the full head rendering, we simply concat the two-stream Gaussians $\{\mu^{h}, c^{h}, R^{h}, S^{h}, \alpha^{h}\}=\text{Concat}(\{\mu^f, c^f, R^f, S^f, \alpha^f\}, \{\mu^e, c^e, R^e, S^e, \alpha^e\})$ and rasterized them into head feature maps:
\begin{align}
\mathcal{M}_h = \mathcal{R}(\{\mu^{h}, c^{h}, R^{h}, S^{h}, \alpha^{h}\}),
\end{align}

\subsection{Expression-Guided Neural Renderer}
To filter noise caused by uneven lighting or camera chromatic aberration~\cite{xu2023gaussianheadavatar}, a UNet-like neural renderer $\mathcal{U}$ opts to rasterize feature maps into the final face-only, eyes, and head images $\{\mathcal{I}_f, \mathcal{I}_e, \mathcal{I}_h\}$:
\begin{equation}
\{\mathcal{I}_f, \mathcal{I}_e, \mathcal{I}_h\} = \mathcal{U}(\{\mathcal{M}_f, \mathcal{M}_e, \mathcal{M}_h\}), 
\end{equation}
However, the vanilla renderer used in Gaussian Head Avatar~\cite{xu2023gaussianheadavatar} and GazeNeRF~\cite{ruzzi2023gazenerf} focuses solely on image post-processing, resulting in a loss of subject-specific facial details.
To further improve generalization ability by providing subject-specific information, we inject the expressive latent codes $\tau$ into the neural renderer through a slice cross-attention module. Let \(z_b\) represent the bottleneck feature obtained from the encoder of \(\mathcal{U}\). We utilize the latent codes to query this bottleneck feature, using it as a conditional signal to guide the renderer's synthesis process. The guiding process can be formulated as:
\vspace{-2mm}
\begin{equation}
z'_b = z_b + z_b \cdot \text{Attn}(q=\tau, k=z_b, v=z_b)),
\end{equation}
where \(\text{Attn}(\cdot)\) denotes the cross-attention operation that fuses the latent codes with the bottleneck feature. Then the refined feature \(z'_b\) is decoded as final images.

\subsection{Training}
\noindent\textbf{GazeGaussian Initialization.}
Initialization for the 3D Gaussians (3DGS) is crucial for stable optimization. Following Gaussian Head Avatar~\cite{xu2023gaussianheadavatar}, we initialize the two-stream Gaussians using the neutral mesh extracted from an SDF field. This neutral mesh provides a coarse geometry and texture, which are used to initialize the positions and features of the Gaussians. To decouple the face-only and eye regions, we compute the 3D neutral landmarks and use learnable parameters to define the vertices near the eyes as the initial Gaussians for the eye region, while the rest of the head is used to initialize the face-only Gaussians. Additionally, we transfer the parameters of all deformation and color MLPs while the MLPs for attribute prediction and the expression-guided neural renderer are randomly initialized.

\noindent\textbf{Image Synthesis Loss.} 
The masked ground truth image $\mathcal{I}_{gt}$ is used to supervise the rendered images ${\mathcal{I}_f, \mathcal{I}_e, \mathcal{I}_h}$, corresponding to the face-only, eyes, and head regions, respectively. Additionally, we enforce the first three channels of the feature maps ${\mathcal{M}_f, \mathcal{M}_e, \mathcal{M}_h}$ to learn RGB colors. For rendered images and the first three channels of feature maps, we apply the same loss functions. For brevity, we take the loss for the rendered eye image $\mathcal{I}_e$ as an example. We mask the ground truth image using an eye mask and then apply L1 loss, SSIM loss, and LPIPS loss on the masked image:
\begin{equation}
\begin{split}
    \mathcal{L}_{\mathcal{I}}^e = ||\mathcal{I}_{gt}-\mathcal{I}_e||_1 &+ \lambda_{SSIM}(1 - SSIM(\mathcal{I}_{gt}, \mathcal{I}_e)) \\ &+ \lambda_{VGG}VGG(\mathcal{I}_{gt}, \mathcal{I}_e),
\end{split}
\end{equation}
where $\lambda_{SSIM}=\lambda_{VGG}=0.1$ is the weight of loss. The image synthesis loss is the sum of the three rendered images and three feature maps:
\begin{equation}
\begin{split}
    \mathcal{L}_{\mathcal{I}} = \mathcal{L}_{\mathcal{I}}^f + \mathcal{L}_{\mathcal{I}}^e + \mathcal{L}_{\mathcal{I}}^h + \mathcal{L}_{\mathcal{M}}^f + \mathcal{L}_{\mathcal{M}}^e + 
    \mathcal{L}_{\mathcal{M}}^h,
\end{split}
\end{equation}
where $\mathcal{L}_{\mathcal{I}}^f, \mathcal{L}_{\mathcal{I}}^e, \mathcal{L}_{\mathcal{I}}^h$ represent the losses for the rendered face-only, eye, and head images, respectively. $\mathcal{L}_{\mathcal{M}}^f, \mathcal{L}_{\mathcal{M}}^e, \mathcal{L}_{\mathcal{M}}^h$ represent the losses for the feature maps corresponding to the face-only, eye, and head regions, respectively. The image synthesis loss ensures the full disentanglement of the eye and the rest of the face. 

\begin{table*}[t]
\caption{Within-dataset comparison: Quantitative results of the \methodname to other SOTA methods on the ETH-XGaze dataset in terms of gaze and head redirection errors in degree, rendered image quality (SSIM, PSNR, LPIPS, FID), identity similarity and rendering FPS.}
\centering
\resizebox{0.85\textwidth}{!}{
\begin{tabularx}{\textwidth}{P{3.2cm} | C{1.33cm} C{1.33cm} C{1.33cm} C{1.33cm} C{1.33cm} C{1.33cm} C{1.33cm} C{1.33cm}}
\toprule
 Method & Gaze$\downarrow$ & Head Pose$\downarrow$ & SSIM$\uparrow$ & PSNR$\uparrow$ & LPIPS$\downarrow$ & FID$\downarrow$ & Identity Similarity$\uparrow$ & FPS$\uparrow$ \\
\midrule
STED & 16.217 & 13.153 & 0.726 & 17.530 & 0.300 & 115.020 & 24.347 & 18 \\
HeadNeRF & 12.117 & 4.275 & 0.720 & 15.298 & 0.294 & 69.487 & 46.126 & 35 \\
GazeNeRF & 6.944 & 3.470 & 0.733 & 15.453 & 0.291 & 81.816 & 45.207 & 46 \\
Gaussian Head Avatar & 30.963 & 13.563 & 0.638 & 12.108 & 0.359 & 74.560 & 27.272 & \textbf{91} \\
\midrule
GazeGaussian (Ours) & \textbf{6.622} & \textbf{2.128} & \textbf{0.823} & \textbf{18.734} & \textbf{0.216} & \textbf{41.972} & \textbf{67.749} & 74 \\
\bottomrule
\end{tabularx}
}
\label{tab:comare_xgaze}
\end{table*}

\noindent\textbf{Gaze Redirection Loss.} 
To improve task-specific performance and eliminate task-relevant inconsistencies between the target image $\mathcal{I}_{gt}$ and the reconstructed head image $\mathcal{I}_{h}$, we adopt the functional loss used in STED~\cite{selflearning} and GazeNeRF~\cite{ruzzi2023gazenerf}. The gaze redirection loss can be formulated as:
\begin{equation}
    \begin{split}
        \mathcal{L}_{\mathcal{G}} (\mathcal{I}_{h}, \mathcal{I}_{gt}) 
        &=  \mathcal{E}_{\text{ang}}(\psi^g(\mathcal{I}_{h}), \psi^g(\mathcal{I}_{gt})),  
        \\    \mathcal{E}_{\text{ang}}(\varphi, \ \hat{\varphi}) &= \arccos  \frac{\varphi \cdot {\hat{\varphi}}}{\|\varphi\| \, \|{\hat{\varphi}}\|} \, ,
    \end{split}
\end{equation}
where $\psi^g(\cdot)$ represents the gaze direction estimated by a pre-trained gaze estimator network, and $\mathcal{E}_{\text{ang}}(\cdot,\cdot)$ represents the angular error function. Our final loss function is:
\begin{equation}
    \mathcal{L}= \lambda_{\mathcal{I}} \mathcal{L}_{\mathcal{I}} + \lambda_{\mathcal{G}} \mathcal{L}_{\mathcal{G}},
\end{equation}
where $\lambda_{\mathcal{I}}=1.0$ and  $\lambda_{\mathcal{G}}=0.1$. GazeGaussian is trained with the final loss until convergence.

\begin{figure*}[htbp]
    \centering
    \includegraphics[width=0.85\textwidth]{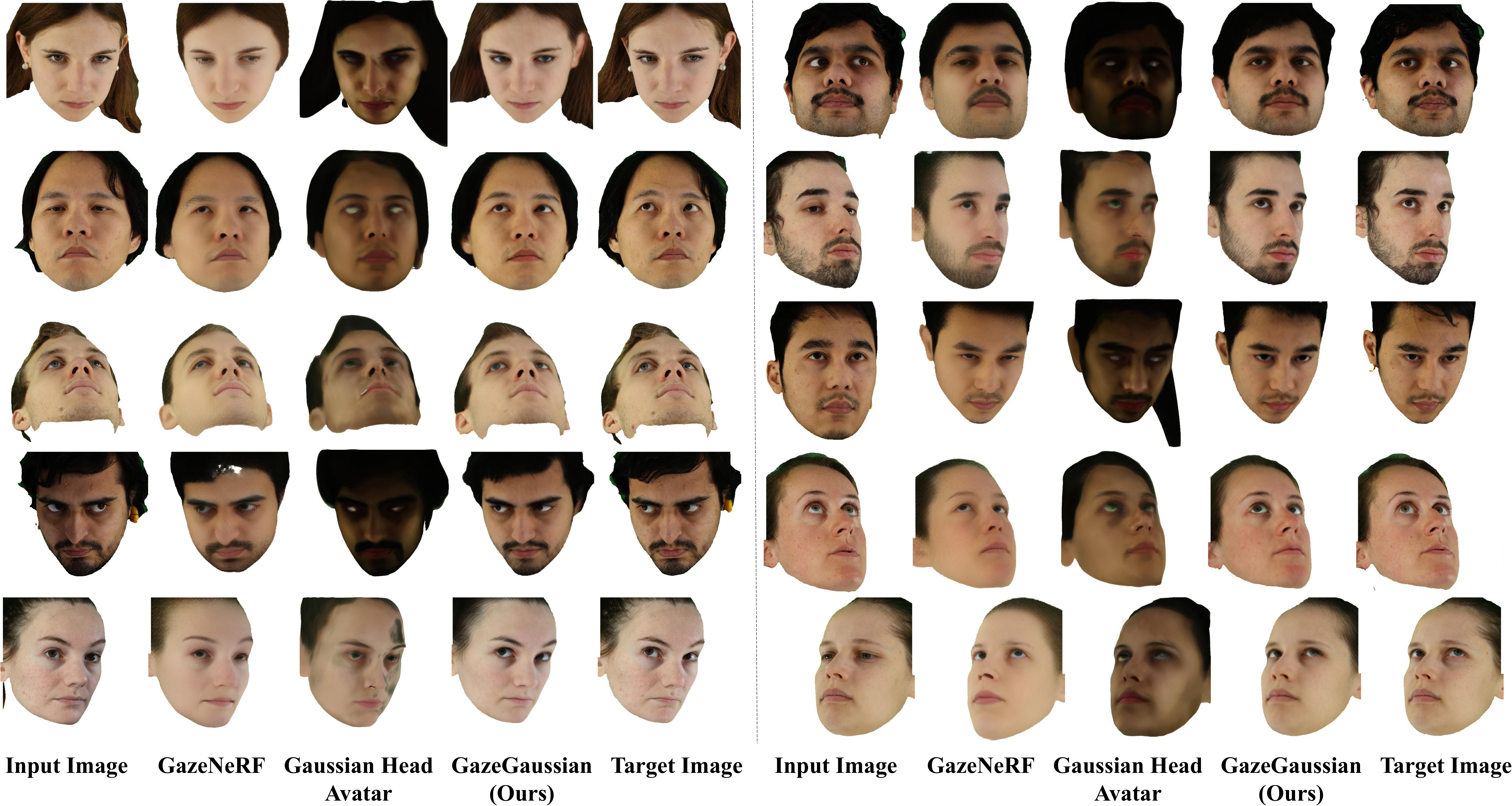}
    \caption{Within-dataset visualization: head images are generated from the ETH-XGaze test set using our GazeGaussian, GazeNeRF, and Gaussian Head Avatar. GazeGaussian generates photo-realistic images with the target gaze direction, preserving identity and facial details. In contrast, GazeNeRF loses identity information and facial details, while Gaussian Head Avatar fails to manipulate the gaze direction.}
    \vspace{-2mm}
    \label{fig:visualization}
\end{figure*}

\begin{figure*}[htbp]
    \centering
    \includegraphics[width=0.95\textwidth]{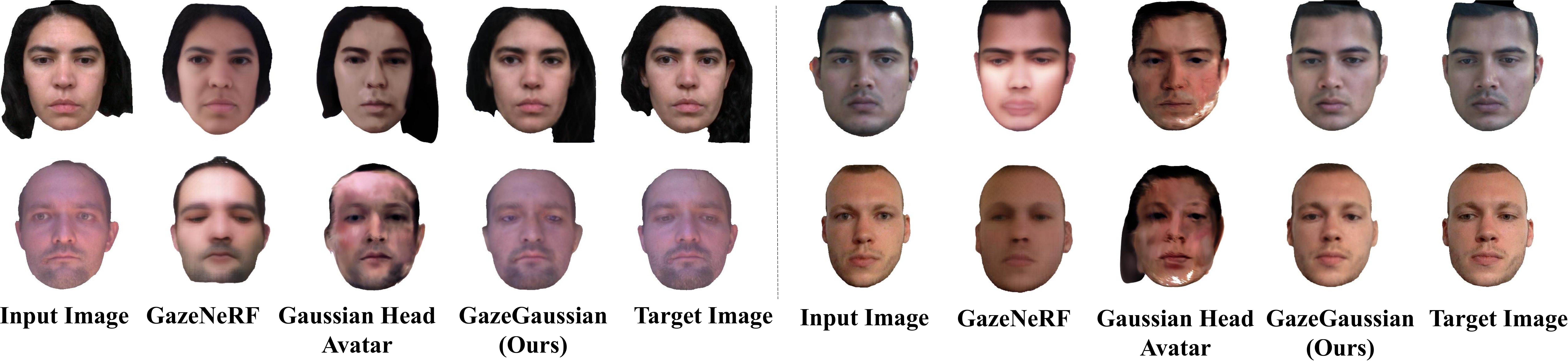}
    \caption{Cross-dataset visualization: head images are from the MPIIFaceGaze test set using our GazeGaussian, GazeNeRF, and Gaussian Head Avatar. Please refer to Sec. \ref{sec: visual_cross_data} in the supplementary for more visualization.}
    \label{fig:visualization_mpii}
    \vspace{-2mm}
\end{figure*}
\section{Experiments}
To demonstrate the effectiveness of \methodname, we first conduct a within-dataset comparison on the ETH-XGaze dataset~\cite{xgaze}, testing GazeGaussian alongside state-of-the-art gaze redirection and head generation methods. Next, we perform a cross-dataset comparison on ColumbiaGaze~\cite{columbia}, MPIIFaceGaze~\cite{mpii, deepappearence}, and GazeCapture~\cite{gazecapture} to assess generalization. We also conduct an ablation study to analyze the contributions of each component in \methodname. All experimental settings strictly follow GazeNeRF~\cite{ruzzi2023gazenerf}, including dataset preprocessing, the ETH-XGaze split (14.4K images for efficient training), image pairs for evaluation, and the same pre-trained gaze estimator. 
Additionally, we validate the impact of synthesized data on gaze estimator performance in Sec.~\ref{sec: gaze_esti_per}. \textbf{Due to space limitations, please refer to Sec.~\ref{sec:suppl_exp} and Sec.~\ref{sec:suppl_vis} in the supplementary for more details on the experiment and visualization results.}
\subsection{Experimental Settings}
\noindent\textbf{Dataset Pre-processing.} Following preprocessing in GazeNeRF~\cite{ruzzi2023gazenerf}, we normalize raw images~\cite{zhang18_etra, sugano2014learning} and resize them into a resolution 512$\times$512. To enable separate rendering of the face and eyes regions, we generate masks using face parsing models~\cite{faceparsingmodel}. We also use the 3D face tracking method from~\cite{xu2023gaussianheadavatar} to produce identity and expression codes and camera poses for the input of our method. For normalization, gaze labels provided by datasets are converted to pitch-yaw angles in the head coordinate system across all datasets. More details are provided in the supplementary materials.

\noindent\textbf{Baselines.} We compare our method with the self-supervised gaze redirection approach STED~\cite{selflearning}, along with NeRF-based models such as HeadNeRF~\cite{headnerf} and the state-of-the-art method GazeNeRF~\cite{ruzzi2023gazenerf}, as well as the latest 3DGS-based head synthesis method, Gaussian Head Avatar~\cite{xu2023gaussianheadavatar}. As the NeRF-based methods, NeRF-Gaze~\cite{yin2024Nerf_gaze} and Wang \etal~\cite{wang2023high} are not yet open-sourced, they are not available for inclusion in our comparisons. 

\noindent\textbf{Metrics.} We evaluate all models using four categories: redirection accuracy, image quality, identity preservation (ID), and rendering speed. Redirection accuracy is measured by gaze and head poses angular errors, using the same ResNet50~\cite{he2016deep}-based estimator in GazeNeRF~\cite{ruzzi2023gazenerf}. Image quality is assessed with SSIM, PSNR, LPIPS, and FID. Identity preservation (ID) is evaluated with FaceX-Zoo~\cite{wang2021facex}, comparing identity consistency between redirected and ground-truth images. Rendering speed is reported as average FPS.
\subsection{Within-dataset Comparison}


Following the experimental setup of GazeNeRF, we perform a within-dataset evaluation to compare the performance of \methodname with other state-of-the-art methods.
All models are trained using 14.4K images derived from 10 frames per subject, with 18 camera view images per frame, covering 80 subjects in the ETH-XGaze training set. For a fair comparison, the training set is strictly the same as GazeNeRF.
Given that Gaussian Head Avatar is not specifically designed for gaze redirection and does not support gaze as a condition, only expressive latent codes are provided for its training and evaluation.
The evaluation is conducted on the person-specific test set of the ETH-XGaze dataset. This test set consists of 15 subjects, each with 200 images annotated with gaze and head pose labels. We follow the pairing setting in GazeNeRF, which pairs these 200 labeled images per subject as input and target samples, and the same pairings are used across all models to ensure fairness. 

Tab.~\ref{tab:comare_xgaze} presents the quantitative results of \methodname alongside baseline methods. 
It can be observed that \methodname consistently outperforms prior methods across all metrics. 
Compared to the previous SOTA method GazeNeRF, which only applies rotation to feature maps for implicit gaze redirection, \methodname adopts a Gaussian eye rotation to explicitly control eye movement. 
Such a technique improves redirection precision, resulting in high-quality head image synthesis.
Additionally, \methodname achieves a rendering speed of 74 FPS, nearly doubling the performance of GazeNeRF, underscoring its efficiency.
In contrast, Gaussian Head Avatar (GHA), the latest 3DGS-based model, fails to achieve high performance in gaze redirection.
The lack of dedicated mechanisms for gaze disentanglement and explicit eye region modeling in GHA leads to poor gaze redirection performance. 
By decoupling the face and eye representation with two-stream Gaussians, \methodname offers both higher accuracy and better visual quality, particularly in challenging scenarios involving extreme head poses or subtle gaze variations.
Furthermore, GHA is limited to animating a single avatar, whereas \methodname with the proposed neural-guided renderer achieves better generalization ability and identity preservation across different subjects. 


We present a visualization comparison of different methods in Fig.~\ref{fig:visualization}. 
GHA fails to generalize to different subjects and struggles to preserve personal identity in the rendered images, which is quantitatively verified as the low `identity similarity' in Tab.~\ref{tab:comare_xgaze}. 
Moreover, GHA produces blurred and unrealistic eye regions, significantly degrading the visual quality of gaze redirection.
GazeNeRF, which implicitly rotates the feature map, fails to effectively control eye appearance under extreme gaze directions (as shown in the last row). Furthermore, GazeNeRF struggles with rendering fine-grained facial details and exhibits notable artifacts in hair rendering.
The failure to precisely render eye details in GHA and GazeNeRF restricts their capability in gaze redirection. In contrast, \methodname consistently produces highly realistic results, even under challenging conditions, setting a new benchmark for gaze redirection tasks.

\begin{table*}[!ht]
\caption{Cross-dataset comparison: Quantitative results of GazeGaussian to other SOTA baselines on ColumbiaGaze, MPIIFaceGaze, and GazeCapture datasets in terms of gaze and head redirection errors in degree, LPIPS, and identity similarity (ID).}
\vspace{-2mm}
\centering
\resizebox{0.9\textwidth}{!}{
\begin{tabularx}{\textwidth}{P{3.2cm} | C{0.7cm} C{0.7cm} C{0.7cm} C{0.85cm} | C{0.7cm} C{0.7cm} C{0.7cm} C{0.85cm} | C{0.7cm} C{0.7cm} C{0.7cm} C{0.85cm}}
\toprule
\multirow{2}{*}{Method} & \multicolumn{4}{c|}{ColumbiaGaze} & \multicolumn{4}{c|}{MPIIFaceGaze} & \multicolumn{4}{c}{GazeCapture} \\
 & Gaze$\downarrow$ & Head$\downarrow$ & LPIPS$\downarrow$ & ID$\uparrow$ & Gaze$\downarrow$ & Head$\downarrow$ & LPIPS$\downarrow$ & ID$\uparrow$ & Gaze$\downarrow$ & Head$\downarrow$ & LPIPS$\downarrow$ & ID$\uparrow$ \\
\midrule

STED & 17.887 & 14.693 & 0.413 & 6.384 & 14.796 & 11.893 & 0.288 & 10.677 & 15.478 & 16.533 & 0.271 & 6.807 \\
HeadNeRF & 15.250 & 6.255 & 0.349 & 23.579 & 14.320 & 9.372 & 0.288 & 31.877 & 12.955 & 10.366 & 0.232 & 20.981 \\
GazeNeRF & 9.464 & 3.811 & 0.352 & 23.157 & 14.933 & 7.118 & 0.272 & 30.981 & 10.463 & 9.064 & 0.232 & 19.025 \\
Gaussian Head Avatar & 10.939 & 3.953 & 0.347 & 46.183 & 12.021 & 8.530 & 0.295 & 30.932 & 11.571 & 7.664 & 0.295 & 22.236 \\
\midrule
GazeGaussian (Ours) & \textbf{7.415} & \textbf{3.332} & \textbf{0.273} & \textbf{59.788} & \textbf{10.943} & \textbf{5.685} & \textbf{0.224} & \textbf{41.505} & \textbf{9.752} & \textbf{7.061} & \textbf{0.209} & \textbf{44.007} \\
\bottomrule
\end{tabularx}
}
\label{tab:comare_otherdataset}
\end{table*}

\begin{table*}[!ht]
\centering
\caption{Component-wise ablation study of \methodname on the ETH-XGaze dataset in terms of gaze and head redirection errors in degree, redirection image quality (SSIM, PSNR, LPIPS and FID), and identity similarity (ID).}
\vspace{-1mm}
\resizebox{0.9\textwidth}{!}{
\begin{tabularx}{\textwidth}{C{1.3cm} C{1.3cm} C{1.3cm} C{1.3cm} C{1.3cm} C{1.3cm} C{1.3cm} C{1.3cm} C{1.3cm} C{1.3cm} }
\toprule
Two-stream & Gaussian Eye Rep. & Expression-Guided & Gaze$\downarrow$ & Head Pose$\downarrow$ & SSIM$\uparrow$ & PSNR$\uparrow$ & LPIPS$\downarrow$ & FID$\downarrow$ & Identity Similarity$\uparrow$\\
\midrule
\checkmark &  &  & 13.651 & 2.981 & 0.753 & 16.376 & 0.272 & 55.481 & 38.941 \\
\checkmark & & \checkmark & 13.489 & 3.149 & 0.751 & 16.365 & 0.274 & 54.327 & 41.521  \\
\checkmark & \checkmark & & 8.883 & 2.635 & 0.766 & 16.692 & 0.254 & 48.891 & 45.013 \\
 & \checkmark & \checkmark & 7.494 & 3.098 & 0.769 & 16.873 & 0.250 & 49.658 & 46.155 \\
\checkmark & \checkmark & \checkmark & \textbf{6.622} & \textbf{2.128} & \textbf{0.823} & \textbf{18.734} & \textbf{0.216} & \textbf{41.972} & \textbf{67.749} \\
\bottomrule
\end{tabularx}
}
\vspace{-2mm}
\label{tab:comare_ablation}
\end{table*}

\subsection{Cross-dataset Comparison}
To access the generalization capability of \methodname, we train all methods on the ETH-Xgaze dataset and evaluate them on three other datasets: ColumbiaGaze, MPIIFaceGaze, and the test set of GazeCapture. The training setup remains consistent with the within-dataset evaluation, using the same model configurations and trained parameters. 

The results shown in Tab.~\ref{tab:comare_otherdataset} and Fig.~\ref{fig:visualization_mpii} demonstrate that \methodname consistently outperforms all other methods across the three datasets and all evaluation metrics. 
By introducing a novel expression-guided neural renderer, \methodname gains better generalization ability by integrating person-specific features, resulting in optimal identity preservation.
On the other hand, GHA's performance is limited by its modeling strategy, showing poor adaptability to unseen datasets. It produces less clear eye regions and achieves significantly lower identity similarity scores compared to \methodname. These results further validate the superiority of \methodname, making it a more robust choice for handling diverse datasets and complex gaze redirection tasks. 

\begin{figure}[!t]
    \centering
    \includegraphics[width=1.0\linewidth]{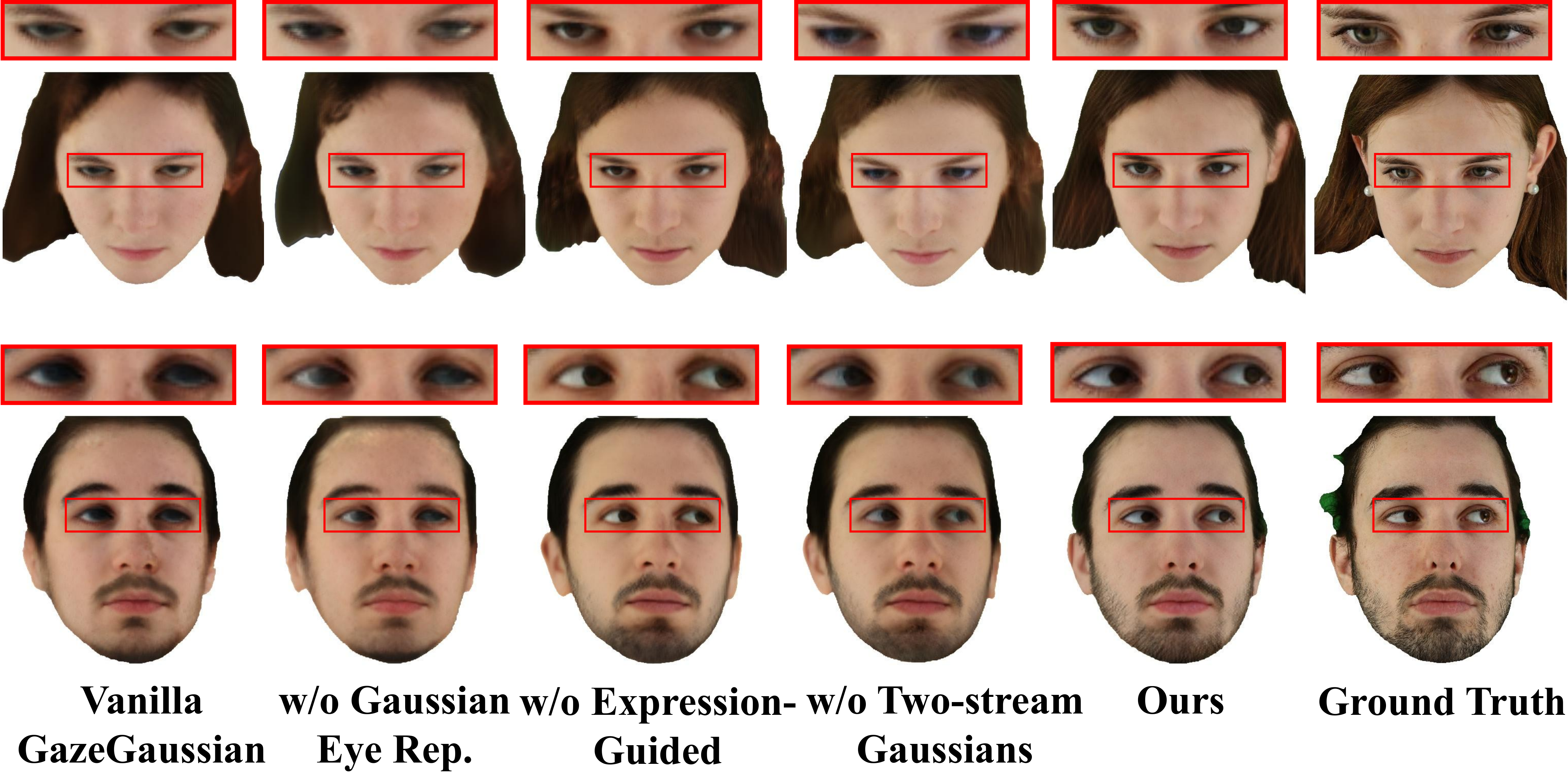}
    \caption{Qualitative ablation study on the ETH-XGaze dataset.}
    \label{fig:ablation_visualization}
    \vspace{-6mm}
\end{figure}

\subsection{Ablation Study}


To validate the effectiveness of each component, we conduct a component-wise ablation study on the ETH-XGaze dataset. 
The results are shown in Tab.~\ref{tab:comare_ablation} and Fig.~\ref{fig:ablation_visualization}. Due to the space limitation, Please refer to Sec.~\ref{sec:suppl_abla_cross} in the supplementary material for the ablation study on cross-dataset. 

\noindent\textbf{Vanilla-GazeGaussian.} In this version, we omit the proposed Gaussian eye rotation representation and expression-guided neural renderer. The corresponding experimental results are shown in the first row of the table and the first column of the visualizations. The eye deformation is treated the same as the face, and the neural renderer remains unchanged from GazeNeRF. The results show that, due to the lack of control over eye rotation, gaze redirection errors are large, and the image synthesis quality is relatively low.
\newline
\noindent\textbf{w/o Gaussian eye rotation representation.} 
To verify the contribution of the proposed Gaussian eye rotation representation, we omit it in the GazeGaussian. The results are shown in the second row of the table and the second column of the figure. Compared to the full version of GazeGaussian, the introduction of a specialized representation for eye deformation significantly improves gaze redirection accuracy and enhances the detail in the eye region.
\newline
\noindent\textbf{w/o Expression-Guided.} We remove the expression-guided neural renderer and rely solely on the vanilla neural renderer in GazeNeRF for image synthesis. The results, shown in the third row of the table and the third column of the figure, indicate a noticeable decline in image quality. Without expression guidance, the model struggles to preserve identity across different subjects, leading to worse identity similarity. The rendered images exhibit lower fidelity in capturing facial details owing to the absence of person-specific features.
\newline
\noindent\textbf{w/o Two-stream.} Replacing the two-stream structure with a single-stream Gaussian model for both face and eye regions leads to performance degradation and loss of synthesis details, as shown in the fourth row of the table and the fourth column of the figure. Combining face and eye regions in a single stream fails to capture the eye region's complex dynamics, resulting in less accurate gaze redirection and lower image fidelity. The two-stream architecture, which decouples the face and eye regions, enables more precise modeling of each region's unique characteristics, improving gaze accuracy and image quality.

Among the ablation experiments, the full \methodname achieves the best performance. This improvement results from the two-stream Gaussian structure, which decouples the face and eye regions, and the proposed Gaussian eye rotation representation, which enables accurate control of eye rotation. Additionally, the expression-guided neural renderer enhances the model's ability to generalize across subjects while preserving facial details. 

\begin{figure}[h!]
\centering
\begin{tikzpicture}[scale=0.85]
\begin{axis}[
    width=10cm, 
    height=8cm, 
    grid=both, 
    grid style={dotted, gray!50}, 
    xlabel={Number of real samples}, 
    ylabel={Error (degrees)}, 
    xmin=1, xmax=9, 
    ymin=5, ymax=10, 
    xtick={1,2,3,4,5,6,7,8,9}, 
    ytick={6,7,8,9,10}, 
    legend style={font=\footnotesize, at={(0.72,0.70)}, anchor=south}, 
    legend cell align={left}, 
    tick label style={font=\small}, 
    xlabel style={font=\small}, 
    ylabel style={font=\small}, 
    mark size=1pt 
]

\addplot[
    color=blue,
    mark=*,
    line width=1pt
] coordinates {
    (1, 9.2)
    (2, 9.0)
    (3, 8.8)
    (4, 7.95)
    (5, 6.8)
    (6, 6.9)
    (7, 6.7)
    (8, 6.75)
    (9, 6.6)
};
\addlegendentry{Real Samples Only}

\addplot[
    color=green,
    mark=*,
    line width=1pt
] coordinates {
    (1, 8.75)
    (2, 8.5)
    (3, 8.95)
    (4, 8.90)
    (5, 7.96)
    (6, 8.2)
    (7, 8.05)
    (8, 8.1)
    (9, 7.90)
};
\addlegendentry{Real + STED}

\addplot[
    color=yellow,
    mark=*,
    line width=1pt
] coordinates {
    (1, 6.95)
    (2, 6.48)
    (3, 6.45)
    (4, 6.05)
    (5, 6.35)
    (6, 6.28)
    (7, 6.15)
    (8, 6.0)
    (9, 5.8)
};
\addlegendentry{Real + GazeNeRF}

\addplot[
    color=red,
    mark=*,
    line width=1pt
] coordinates {
    (1, 6.10)
    (2, 5.96)
    (3, 5.91)
    (4, 5.78)
    (5, 5.74)
    (6, 5.60)
    (7, 5.42)
    (8, 5.31)
    (9, 5.26)
};
\addlegendentry{Real + Ours GazeGaussian}

\end{axis}
\end{tikzpicture}
\vspace{-2mm}
\caption{Improvement for gaze estimation accuracy.}
\label{fig:personal_comparison}
\vspace{-4mm}
\end{figure}
\subsection{Improving gaze estimation performance}
\label{sec: gaze_esti_per}
Following GazeNeRF, we evaluate the improvement of each method for downstream gaze estimation tasks.
Specifically, following settings in GazeNeRF, given a few calibration samples from person-specific ETH-Xgaze test sets, we augment these real samples with gaze-redirected samples generated by GazeGaussian. We then fine-tune the gaze estimator (same in GazeNeRF) pre-trained on ETH-XGaze using these augmented samples and compare its performance with a baseline model fine-tuned only on real samples. 
To ensure a fair comparison, the total number of augmented samples is fixed at 200 (real + generated samples), and we vary the number of real samples used for fine-tuning during the evaluation.

As shown in Fig.~\ref{fig:personal_comparison}, the x-axis represents the number of real samples used, and the y-axis shows the gaze estimation error in degrees on the ETH-XGaze person-specific test set. We evaluate up to nine real samples in the few-shot setting. Fine-tuning the pre-trained gaze estimator with real and generated samples from GazeGaussian achieves the lowest gaze estimation error across all settings. In contrast, samples generated by GazeNeRF and STED lead to higher gaze estimation errors. 
The results clearly demonstrate that the generated samples from GazeGaussian are of higher fidelity and more effective for improving gaze estimation accuracy. 
\vspace{-3mm}
\section{Conclusion}
\vspace{-1mm}
We present GazeGaussian, the first high-fidelity gaze redirection pipeline that uses a two-stream 3DGS model for face and eye disentanglement, along with an expression-conditional neural renderer. 
Numerous experiments have demonstrated that GazeGaussian achieves state-of-the-art performance on the task of gaze direction, paving the way for more robust gaze estimation in real-world applications.

\paragraph{Acknowledgments.} This work was supported by the Beijing Natural Science Foundation (L247008), the NSFC (62332015), the project of China Disabled Persons Federation (CDPF2023KF00002), and the Project of ISCAS (ISCAS-ZD-202401,  ISCAS-JCMS-202401). 

{
    \small
    \bibliographystyle{ieeenat_fullname}
    \bibliography{main}
}
\clearpage
\setcounter{page}{1}
\maketitlesupplementary

\begin{table*}[!ht]
\caption{Component-wise ablation study of \methodname on the ColumbiaGaze, MPIIFaceGaze and GazeCapture datasets.}
\vspace{-3mm}
\centering
\small
\resizebox{0.88\textwidth}{!}{
\begin{tabularx}{\textwidth}{C{0.7cm} C{0.9cm} C{0.9cm} | C{0.7cm} C{0.7cm} C{0.7cm} C{0.8cm} | C{0.7cm} C{0.7cm} C{0.7cm} C{0.8cm} | C{0.7cm} C{0.7cm} C{0.7cm} C{0.8cm}}
\toprule
\multirow{2}{*}{\shortstack{Two-\\stream}} & \multirow{2}{*}{\shortstack{Gaus.\\Eye Rep.}} & \multirow{2}{*}{\shortstack{Exp.\\Guided}} & \multicolumn{4}{c|}{ColumbiaGaze} & \multicolumn{4}{c|}{MPIIFaceGaze} & \multicolumn{4}{c}{GazeCapture} \\
  &  &  & Gaze$\downarrow$ & Head$\downarrow$ & LPIPS$\downarrow$ & ID$\uparrow$ & Gaze$\downarrow$ & Head$\downarrow$ & LPIPS$\downarrow$ & ID$\uparrow$ & Gaze$\downarrow$ & Head$\downarrow$ & LPIPS$\downarrow$ & ID$\uparrow$ \\
\midrule

\checkmark &  &             & 8.996 & 4.494 & 0.325 & 49.286 & 19.787 & 8.491 & 0.321 & 34.483 & 15.697 & 13.740 & 0.260 & 33.393 \\
\checkmark & & \checkmark   & 9.143 & 4.509 & 0.324 & 52.805 & 16.689 & 8.578 & 0.303 & 35.194 & 15.926 & 14.869 & 0.261 & 36.004 \\
\checkmark & \checkmark &   & 7.799 & 3.754 & 0.284 & 57.252 & 11.938 & 6.860 & 0.257 & 35.614 & 10.339 & 8.208 & 0.216 & 40.458 \\
 & \checkmark & \checkmark  & 7.710 & 3.899 & 0.280 & 58.969 & 12.559 & 6.188 & 0.246 & 37.444 & 11.296 & 8.460 & 0.224 & 42.294 \\
\checkmark & \checkmark & \checkmark & \textbf{7.415} & \textbf{3.332} & \textbf{0.273} & \textbf{59.788} & \textbf{10.943} & \textbf{5.685} & \textbf{0.224} & \textbf{41.505} & \textbf{9.752} & \textbf{7.061} & \textbf{0.209} & \textbf{44.007} \\

\bottomrule
\end{tabularx}
}
\label{tab:comare_ablation_cross}
\vspace{-4mm}
\end{table*}

\begin{table*}[t]
\small
\caption{Comparison between baselines + expression-guided neural renderer and GazeGaussian on ETH-xgaze}
\vspace{-3mm}
\centering
\resizebox{0.88\textwidth}{!}{
\begin{tabularx}{\textwidth}{P{3.2cm} | C{1.33cm} C{1.33cm} C{1.33cm} C{1.33cm} C{1.33cm} C{1.33cm} C{1.33cm} C{1.33cm}}
\toprule
 Method & Gaze$\downarrow$ & Head Pose$\downarrow$ & SSIM$\uparrow$ & PSNR$\uparrow$ & LPIPS$\downarrow$ & FID$\downarrow$ & Identity Similarity$\uparrow$ & FPS$\uparrow$ \\
\midrule
GazeNeRF                    & 6.944 & 3.470 & 0.733 & 15.453 & 0.291 & 81.816 & 45.207 & 46 \\
GazeNeRF + EGNR             & 6.854 & 3.025 & 0.764 & 16.147 & 0.258 & 67.219 & 50.268 & 44 \\
GHA        & 30.963 & 8.498 & 0.638 & 12.108 & 0.359 & 74.560 & 27.272 & \textbf{91} \\
GHA + EGNR & 28.374 & 6.533 & 0.714 & 14.213 & 0.305 & 69.101 & 41.332 & 90 \\
\midrule
GazeGaussian (Ours) & \textbf{6.622} & \textbf{2.128} & \textbf{0.823} & \textbf{18.734} & \textbf{0.216} & \textbf{41.972} & \textbf{67.749} & 74 \\
\bottomrule
\end{tabularx}
}
\label{tab:comare_gazenerf_expre_xgaze}
\vspace{-5mm}
\end{table*}

\section{Overview}
\vspace{-2mm}
The supplementary material encompasses the subsequent components. 

\begin{itemize}
    \item Video for continuous gaze redirection
    \item Implementation details
    \item Dataset and pre-processing details
    \item Supplementary experiments
    \begin{itemize}
        \item[--] Ablation study on cross-dataset
        \item[--] Comparison with the FLAME-based method
        \item[--] GazeGaussian vs baseline + expression-guided
    \end{itemize}
    \item Additional visualization results
    \begin{itemize}
        \item[--] Visualization for transformed Gaussians
        \item[--] Visualization for identity morphing
        \item[--] Visualization for ablation study
        \item[--] Visualization for cross-dataset comparison
    \end{itemize}
    \item Ethical considerations and limitations
\end{itemize}
\section{Video for continuous gaze redirection}
Please refer to the video \textbf{``continuous gaze redirection.mp4''} in the supplementary material for continuous gaze redirection results on the ETH-Xgaze. The side-by-side visualization showcases smooth transitions and high-quality novel gaze synthesis produced by GazeGaussian.
\section{Implementation details}
We use the Adam optimizer~\cite{adam}, with a learning rate that follows an exponential decay schedule, starting at $1 \times 10^{-4}$.
We use the VGG-based network pre-trained on ImageNet, as provided by the GazeNeRF~\cite{ruzzi2023gazenerf} implementation, and fine-tune it on the ETH-XGaze training set for the functional loss $\mathcal{L}_{\mathcal{G}}$ as the pre-trained gaze estimator. Additionally, we utilize the ResNet50 backbone from the GazeNeRF~\cite{ruzzi2023gazenerf} framework, trained on the ETH-XGaze training set, to output gaze and head pose for evaluation purposes.
All experiments are conducted on an NVIDIA 4090 GPU. We first train an SDF network to extract the neutral mesh and initialize the two-stream Gaussian parameters in 10 epochs. The full pipeline is then trained for an additional 20 epochs until convergence. The loss weights are described in the method section.
\section{Dataset and pre-processing details}
Following the baseline GazeNeRF~\cite{ruzzi2023gazenerf}, all experiments are conducted on four widely used datasets.
\newline
\textbf{ETH-XGaze}~\cite{xgaze} is a large-scale gaze estimation dataset featuring high-resolution images across a wide range of head poses and gaze directions. Captured with a multi-view camera setup under varying lighting conditions, it includes 756,000 frames from 80 subjects for training. Each frame contains images from 18 different camera perspectives. Additionally, a person-specific test set includes 15 subjects, each with 200 images provided with ground-truth gaze data.
\newline
\textbf{ColumbiaGaze}~\cite{columbia} contains 5,880 high-resolution images from 56 subjects. For each subject, images were taken in five distinct head poses, with each pose covering 21 preset gaze directions, allowing for detailed gaze estimation in controlled conditions.
\newline
\textbf{MPIIFaceGaze}~\cite{mpii, deepappearence} is tailored for appearance-based gaze prediction. MPIIFaceGaze offers 3,000 face images for each of 15 subjects, paired with two-dimensional gaze labels to facilitate gaze estimation research.
\newline
\textbf{GazeCapture}~\cite{gazecapture} is a large-scale dataset collected through crowd-sourcing, featuring images captured across different poses and lighting conditions. For cross-dataset comparison, we use only the test portion, which includes data from 150 distinct subjects.
\newline
\textbf{Pre-processing.} We follow the preprocessing steps in GazeNeRF~\cite{ruzzi2023gazenerf} and Gaussian Head Avatar~\cite{xu2023gaussianheadavatar}. The original resolution of ETH-XGaze~\cite{xgaze} images is 6K × 4K, while images from other datasets vary in resolution. To standardize, we preprocess all images using the normalization method, aligning the rotation and translation between the camera and face coordinate systems. The normalized distance from the camera to the face center is fixed at 680mm. To extract 3DMM parameters and generate masks for the eyes and face-only regions, we utilize the face parsing model from~\cite{faceparsingmodel}. GazeGaussian is trained on a single NVIDIA 4090 GPU for 20 epochs on the train set from ETH-XGaze. During inference, GazeGaussian fine-tunes on a single input image, taking approximately 30 seconds for fine-tuning and 0.2 seconds per image for generation.

\section{Supplementary experiments}
\label{sec:suppl_exp}

\subsection{Ablation study on cross-dataset}
\label{sec:suppl_abla_cross}
To further validate the effectiveness of each proposed component, we conduct an ablation study on the cross-dataset evaluation to assess the generalization capability of our full pipeline. As shown in Tab.~\ref{tab:comare_ablation_cross}, the results are consistent with the ablation study in the main text. The proposed Gaussian eye rotation representation significantly improves eye redirection accuracy while ensuring robust redirection across cross-domain datasets. Additionally, the expression-guided neural renderer preserves the identity characteristics of the input image, enabling generalization ability across different subjects. From the ablation study on cross-dataset, we can further validate the importance of each component.
\subsection{Comparison with the FLAME-based method}
\setlength{\intextsep}{4pt} 
\setlength{\columnsep}{8pt} 
\begin{wrapfigure}{r}{0.6\linewidth}
\centering
\captionof{table}{Image quality comparison with the FLAME-based method.}
\resizebox{1.0\linewidth}{!}{%
\begin{tabular}{l | c | c | c}
    \hline
    Methods & SSIM~$\uparrow$ & PSNR~$\uparrow$ & LPIPS~$\downarrow$ \\
    \hline
    Wang et al. & 0.732 & 19.144 & 0.265 \\
    \hline
    \textbf{Ours} & \textbf{0.823} & 18.734 & \textbf{0.216} \\
    \hline
\end{tabular}%
}

\label{table:FLAME_based_quality}

\end{wrapfigure}
The FLAME-based baseline by Wang et al.~\cite{wang2023high} is not open source and lacks metrics in gaze redirection in its published materials.
Nonetheless, we have cited the reported results on the ETH-Xgaze dataset in Wang's paper and provided a comparison in Tab~\ref{table:FLAME_based_quality}. The results demonstrate that our GazeGaussian still achieves better synthesis quality, especially for perceptual metrics.
\subsection{GazeGaussian vs baseline + expression-guided}
We make a comparison between GazeGaussian and baselines (GazeNeRF, Gaussian Head Avatar) enhanced with the expression-guided neural renderer (EGNR) on the ETH-XGaze dataset. As shown in Tab.~\ref{tab:comare_gazenerf_expre_xgaze}, integrating EGNR into GazeNeRF and Gaussian Head Avatar (GHA) leads to noticeable improvements in gaze redirection accuracy and image quality. This demonstrates the versatility of the proposed expression-guided neural renderer in enhancing facial synthesis and better capturing identity-specific expressions. Although GHA is restricted to animating single head avatar, it can benefit from enhanced generalization through expression-guided neural renderer.
However, even with the added EGNR, the performance of GazeNeRF and GHA remains limited compared to GazeGaussian. The fundamental constraint lies in GazeNeRF’s representation, which lacks the explicit modeling of gaze and facial expression dynamics offered by GazeGaussian’s two-stream Gaussian structure. GHA restricts to full head animation, missing two-stream modeling for face and eye disentanglement, leading to decreased performance. 

\section{Supplementary visualization}
\label{sec:suppl_vis}
\begin{figure}[!ht]
    \centering
    \includegraphics[width=0.95\linewidth]{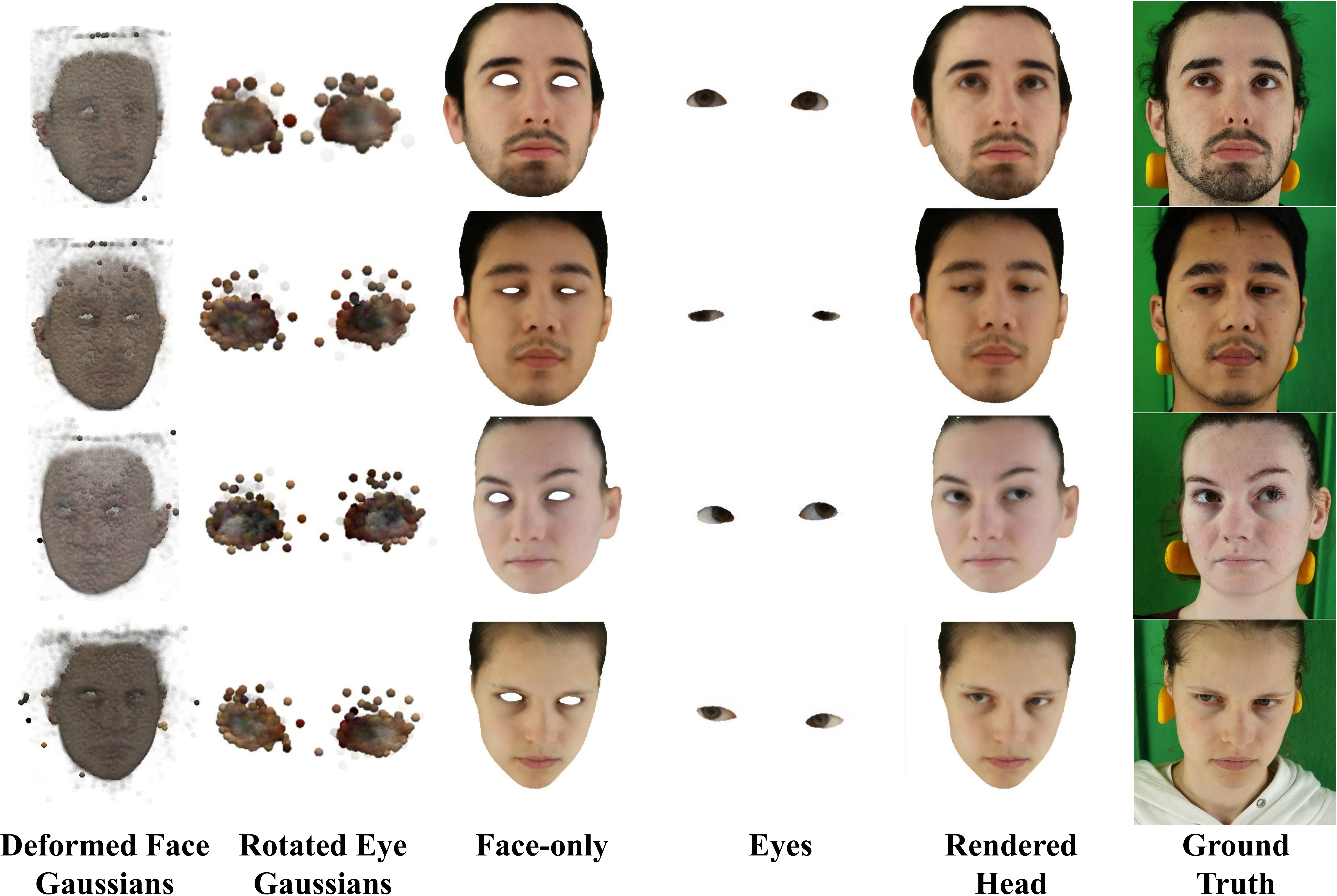}
    \caption{Visualization of transformed two-stream Gaussians after deformation from the canonical space.}
    \label{fig:transformed}
\end{figure}
\subsection{Visualization for transformed Gaussians}
To demonstrate the advantages of GazeGaussian’s explicit control of head pose and gaze direction for head and eye regions, we visualize the Gaussians after deformation from the canonical space. As shown in Fig.~\ref{fig:transformed}, the explicit support for rotation and translation in GazeGaussian allows the deformed Gaussians to form a reasonable spatial distribution and accurate color representation. This capability enables precise geometric control and high-fidelity image rendering. In contrast, GazeNeRF performs rotations only on the feature map level, failing to fully deform in 3D space, which limits its performance compared to our method.

\begin{figure}[htbp]
    \centering
    \includegraphics[width=1.0\linewidth]{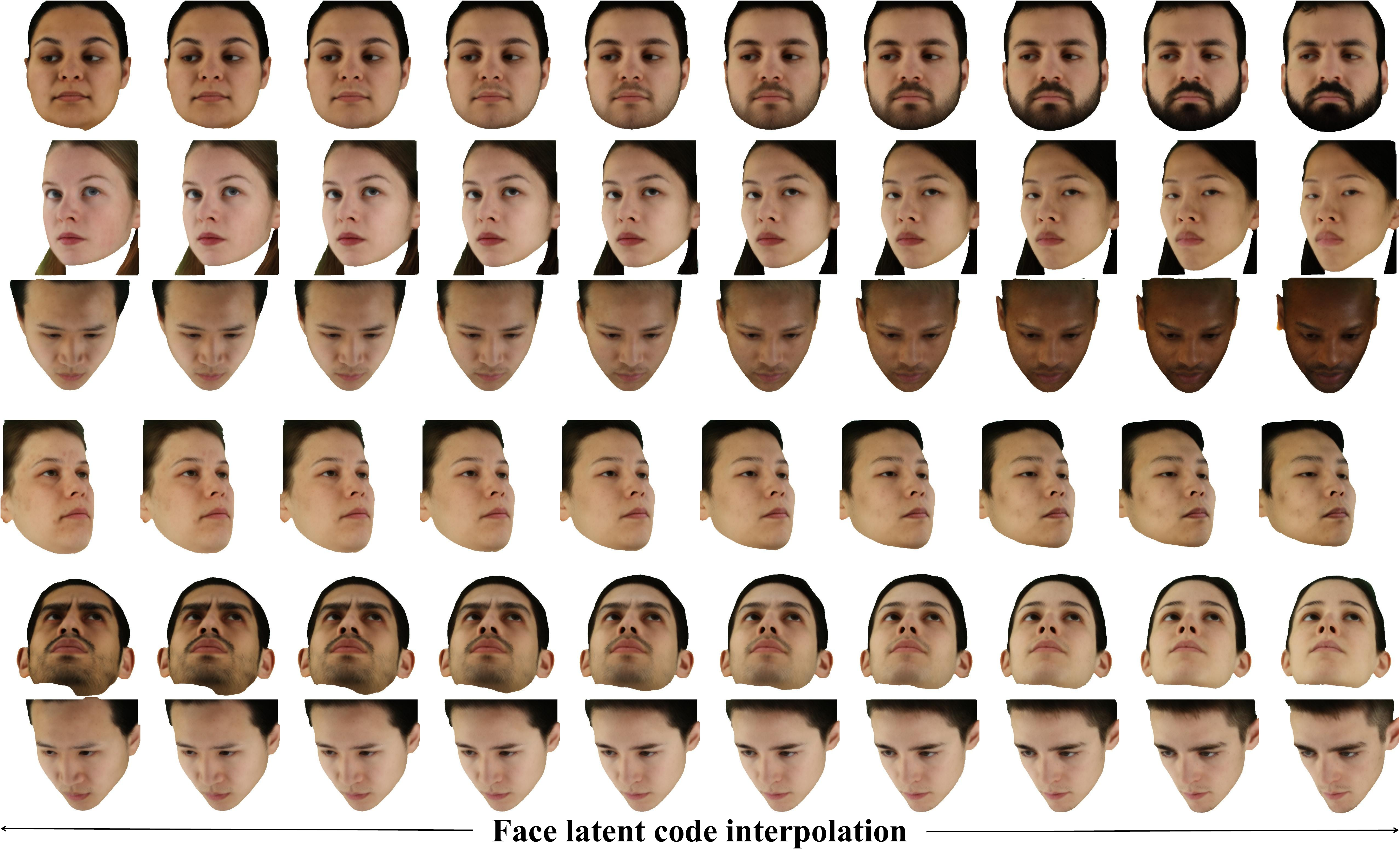}
    \caption{Face morphing results on the ETH-XGaze dataset.}
    \label{fig:face_morphing}
\end{figure}
\subsection{Visualization for identity morphing}
Fig.~\ref{fig:face_morphing} showcases identity morphing results on the ETH-XGaze dataset. We randomly select two subjects with identical gaze directions and head poses. By interpolating their latent codes, we generate a smooth transition between the two identities while keeping the gaze direction and head pose consistent. This visualization demonstrates the capability of GazeGaussian to preserve gaze alignment and head orientation during synthesis, even as the facial features gradually change according to the interpolated latent codes. 

\subsection{Visualization for ablation study}

\begin{figure*}[htbp]
    \centering
    \includegraphics[width=0.97\textwidth]{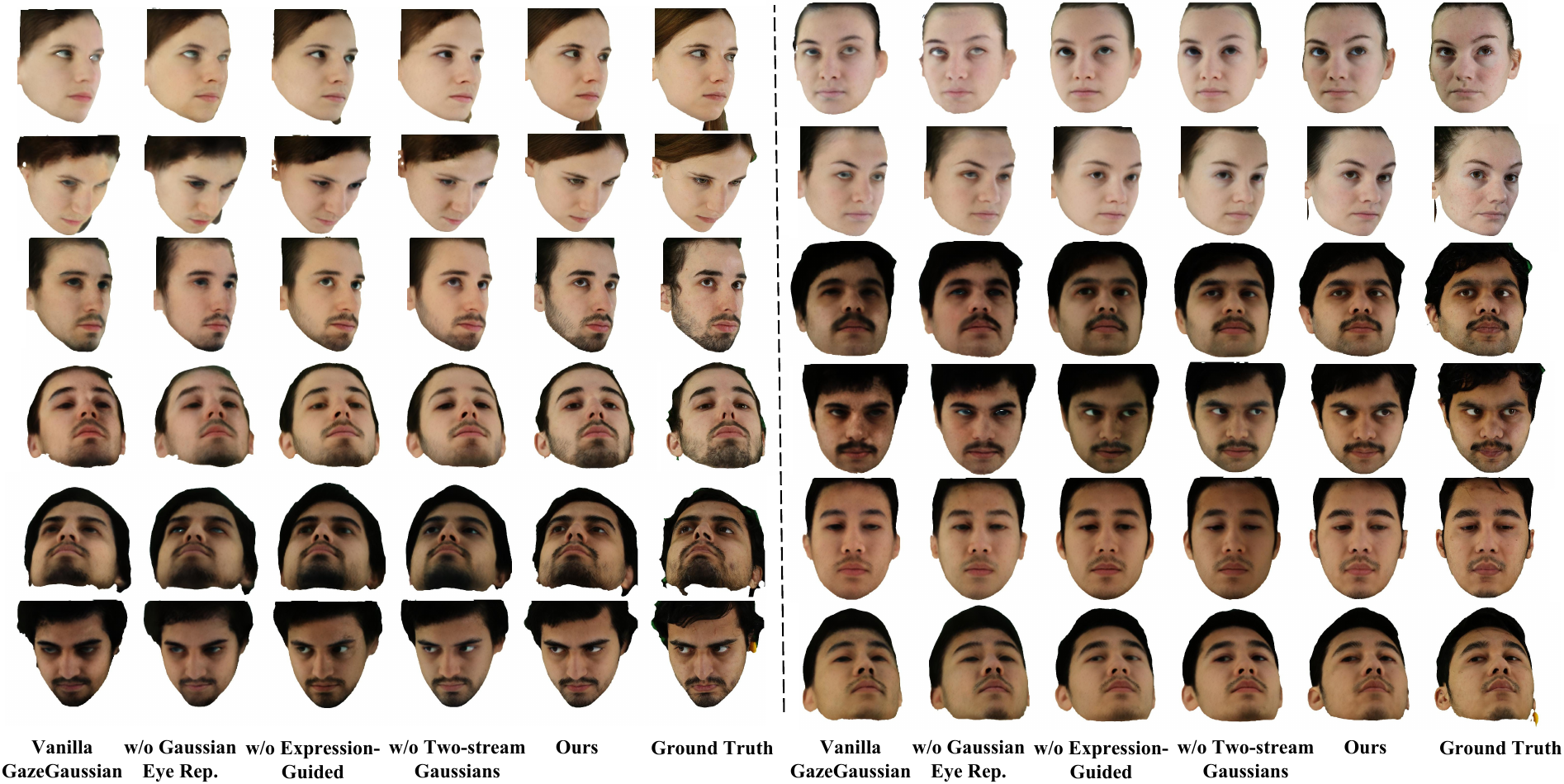}
    \vspace{-2mm}
    \caption{Additional qualitative ablation study on the ETH-XGaze dataset.}
    \label{fig:ablation_visualization_implement}
    \vspace{-4mm}
\end{figure*}

Fig.~\ref{fig:ablation_visualization_implement} presents additional qualitative results from our ablation study conducted on the ETH-XGaze dataset. These visualizations highlight the importance of each proposed component in GazeGaussian.

Without the Gaussian eye rotation representation, the model struggles to achieve accurate eye control, resulting in noticeable deviations in gaze direction and reduced realism in the eye region. This demonstrates the critical role of the Gaussian eye rotation representation in enabling precise and realistic gaze redirection. Additionally, the absence of the expression-guided neural renderer leads to a significant loss in facial detail and expression fidelity. With the renderer included, the synthesized images exhibit finer facial details and improved consistency with the target identity, showcasing the renderer’s effectiveness in enhancing the overall quality of face synthesis. These results confirm that both components contribute significantly to the superior performance and visual fidelity of GazeGaussian.

\subsection{Visualization for cross-dataset comparison}
\label{sec: visual_cross_data}
\begin{figure*}[htbp]
    \centering
    \includegraphics[width=0.97\textwidth]{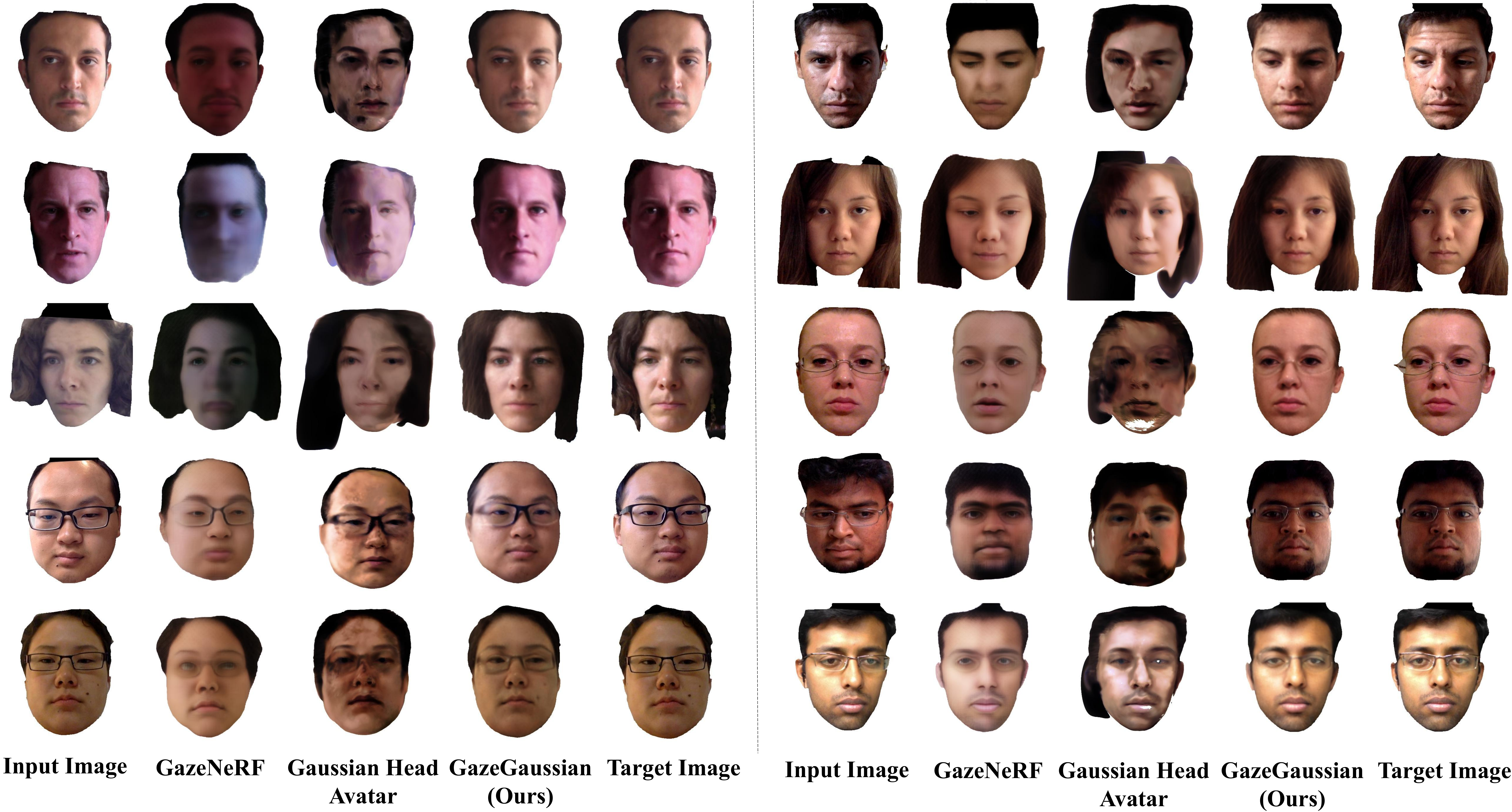}
    \vspace{-2mm}
    \caption{Cross-dataset visualization on MPIIFaceGaze.}
    \label{fig:visualization_mpii_implement}
    \vspace{-4mm}
\end{figure*}

\begin{figure*}[htbp]
    \centering
    \includegraphics[width=0.97\textwidth]{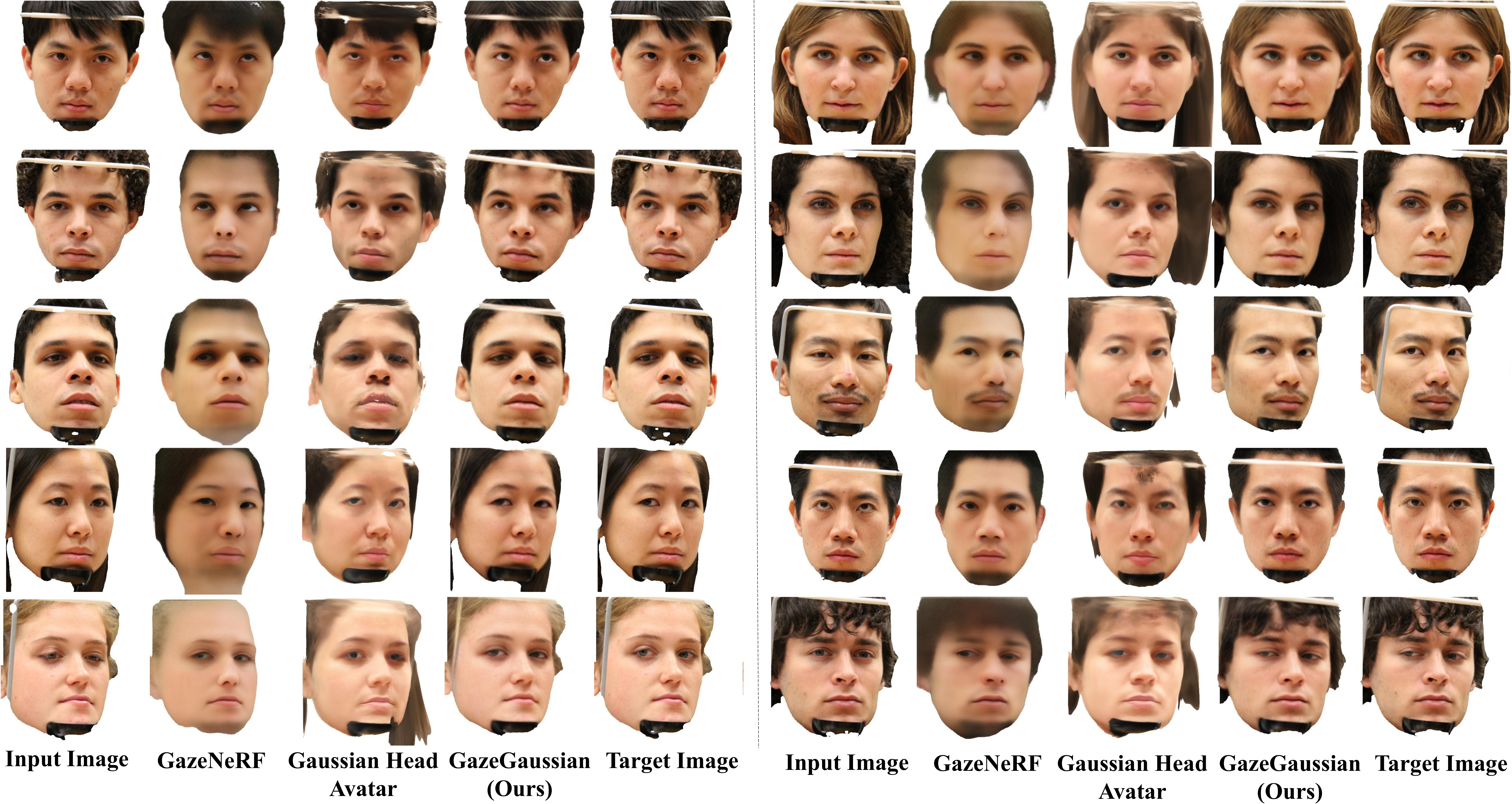}
    \vspace{-2mm}
    \caption{Cross-dataset visualization on ColumbiaGaze.}
    \label{fig:visualization_columbia_implement}
    \vspace{-4mm}
\end{figure*}

\begin{figure*}[htbp]
    \centering
    \includegraphics[width=0.97\textwidth]{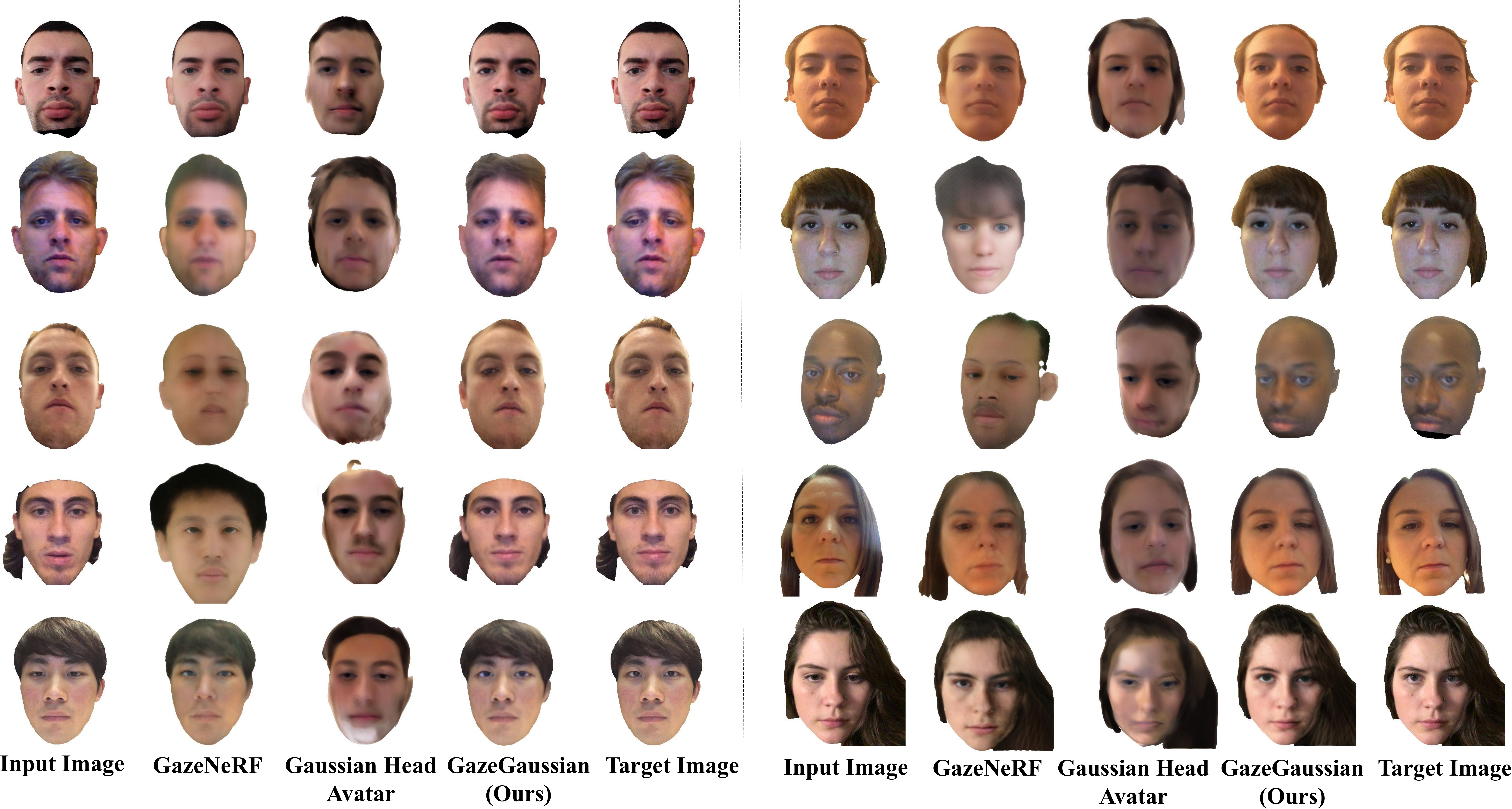}
    \vspace{-2mm}
    \caption{Cross-dataset visualization on GazeCapture. }
    \label{fig:visualization_gazecapture_implement}
    \vspace{-4mm}
\end{figure*}

We provide additional cross-dataset comparison visualizations for MPIIFaceGaze (Fig.~\ref{fig:visualization_mpii_implement}), ColumbiaGaze (Fig.~\ref{fig:visualization_columbia_implement}) and GazeCapture (Fig.~\ref{fig:visualization_gazecapture_implement}). Compared to the baseline, GazeGaussian achieves high-fidelity gaze redirection with superior image synthesis quality.
\section{Ethical considerations and limitations}
\begin{figure}[!ht]
    \centering
    \includegraphics[width=0.75\linewidth]{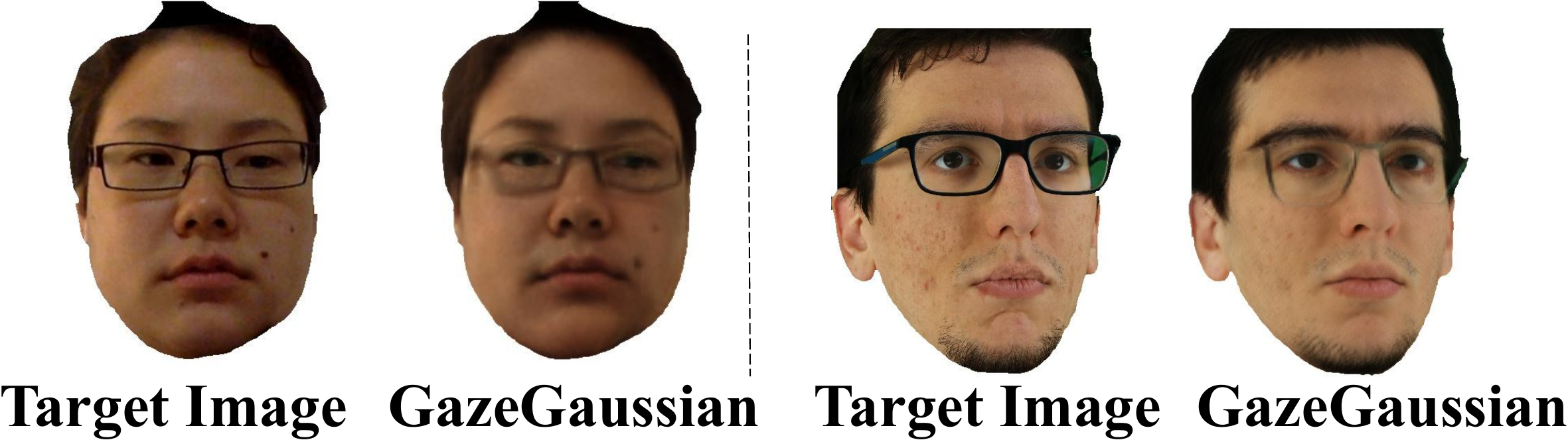}
    \caption{Example of a failure case.}
    \label{fig:fail_case}
\end{figure}
Our approach allows for the creation of lifelike portrait videos that may be exploited to spread misinformation, sway public opinion, and erode trust in media, with grave societal impacts. Thus, developing trustworthy techniques to discern real from fake content is crucial. We firmly oppose any unauthorized or harmful use of this technology and highlight the need to address ethical issues in its implementation.

While GazeGaussian represents a significant advancement in gaze redirection quality,  there is still one unresolved issue. Due to limitations in facial tracking models such as FLAME, it remains challenging to accurately model accessories such as glasses, earrings, and even hair details as shown in Fig.~\ref{fig:fail_case}. An existing method~\cite{luo2024gaussianhair} has attempted to use cylindrical Gaussian representations to model hair. To further enhance GazeGaussian, improving the 3DGS facial representation will be a key focus of our future work.

\end{document}